\documentclass[runningheads]{llncs}
\usepackage[T1]{fontenc}
\usepackage{graphicx}
\usepackage{multirow}
\usepackage{algorithm}
\usepackage{algpseudocode}
\usepackage{amsmath}
\usepackage{amssymb}
\usepackage[font=small, skip=6pt]{caption}
\usepackage{subcaption}

\begin{document}
\title{Safer Skin Lesion Classification with Global Class Activation Probability Map Evaluation and SafeML}
%
%
\author{Kuniko Paxton\inst{1}\orcidID{0009-0000-8897-9775} \and
Koorosh Aslansefat\inst{1}\orcidID{0000-0001-9318-8177} \and
Amila Akagić\inst{2}\orcidID{0000-0002-4795-5424} \and
Dhavalkumar Thakker\inst{1}\orcidID{0000-0003-4479-3500} \and
Yiannis Papadopoulos\inst{1}\orcidID{0000-0001-7007-5153}
}
\authorrunning{K.Paxton et al.}
%
\institute{School of Computer Science, University of Hull, Cottingham Road, Hull, HU6 7RX, United Kingdom \and Faculty of Electrical Engineering, University of Sarajevo, Zmaja od Bosne bb, Sarajevo, 71000, Bosnia and Herzegovina}
\maketitle              
\begin{abstract}
Recent advancements in skin lesion classification models have significantly improved accuracy, with some models even surpassing dermatologists' diagnostic performance. However, in medical practice, distrust in AI models remains a challenge. Beyond high accuracy, trustworthy, explainable diagnoses are essential. Existing explainability methods have reliability issues, with LIME-based methods suffering from inconsistency, while CAM-based methods failing to consider all classes. To address these limitations, we propose Global Class Activation Probabilistic Map Evaluation, a method that analyses all classes' activation probability maps probabilistically and at a pixel level. By visualizing the diagnostic process in a unified manner, it helps reduce the risk of misdiagnosis. Furthermore, the application of SafeML enhances the detection of false diagnoses and issues warnings to doctors and patients as needed, improving diagnostic reliability and ultimately patient safety. We evaluated our method using the ISIC datasets with MobileNetV2 and Vision Transformers. Our code for the experiment is available on GitHub (https://github.com/Kuniko925/ExplainForSafe).
\keywords{SafeML \and Class Activation Map \and Explainability \and Skin lesion classification \and Medical image classification}
\end{abstract}
\section{Introduction}
In recent years,  skin lesion classification models have achieved remarkable improvements in diagnostic accuracy, with some surpassing the performance of dermatologists \cite{esteva2017dermatologist}. These high-performance models have the potential to support clinical decision-making and alleviate physicians' workload. However, despite their effectiveness, studies have consistently reported a lack of trust in such models among medical professionals \cite{metta2021explainable}. To foster wider clinical adoption, it is essential not only to achieve high predictive accuracy but also to enhance model transparency by clearly explaining the basis of predictions  \cite{gertych2024ai,chanda2024dermatologist}.

To this end, various explainability techniques have been proposed, such as Class Activation Mapping (CAM) \cite{zhou2016learning} and local interpretable model-agnostic explanations (LIME) \cite{ribeiro2016should}, to visualize the model's reasoning process. These visualizations have been shown to improve clinicians' trust. However, the reliability of these explanations remains uncertain. While they can expose instances of shortcut learning, the quantitative relationship between the quality of explanations and model performance is still unclear. 

LIME, for example, offers local explanations, but suffers from inconsistency due to sensitivity to kernel size and local perturbations \cite{saadatfar2024us}  \cite{xiang2023stable}. Some efforts have been made to address these issues, but a definite solution remains elusive. CAM-based techniques such as Gradient-weighted Class Activation Mapping (Grad-CAM) \cite{selvaraju2017grad} attempt to highlight visually salient image regions by computing gradients for filters or vectors in the target layer. Variants such as Integrated Gradients \cite{sundararajan2017axiomatic}, seCAM\cite{cao2023novel} and Score-CAM \cite{wang2020score} also emphasize class-relevant features. However, in most studies, the results yielded by application of these CAM tools are used merely as supplementary diagnostics  \cite{hauser2022explainable}. 

We argue that explainability can and should play a more central role, especially in mitigating misdiagnoses and alerting users to anomalous predictions. Most explainability tools only highlight the saliency of the input image with respect to the predicted class  \cite{selvaraju2017grad,sundararajan2017axiomatic,wang2020score}, overlooking insights from all potential classes. This narrow focus can result in misleading visualizations.  For example, even when a model misclassifies a lesion,  as shown in  Fig. \ref{fig:cam_each_class}, if the highlighted region aligns with the lesion area and attention to other classes is absent, the prediction may falsely appear reliable. Such cases present serious risks of misdiagnosis.

To address this, we propose “Global Class Activation Probabilistic Mapping (GCAPM)”, an evaluation method that analyzes activation regions across all classes for a given input. By normalizing and probabilistically aggregating class activation maps on a pixel level, we can more comprehensively assess the model's diagnostic behaviors. Additionally,  class output weights are integrated into visualisation, offering a unified view of attention distribution across all classes, regardless of whether the final prediction is correct. GCAPM produces a segmented output of the model's regions of interest, which can be quantitatively compared to the ground truth lesion annotations using metrics such as sensitivity and false positive rates.

Beyond clinical settings, skin lesion models are increasingly being deployed in mobile applications and web-based self-diagnosis tools, especially in areas with limited access to medical care and long waiting times. These tools, while valuable, may cause potential patients to neglect their illness or place too much trust in the application. To improve the safety of such applications, we incorporate the principles of safeML as proposed in \cite{aslansefat2020safeml}. SafeML monitors data drift of operational data from training data using Empirical Cumulative Distribution Functions (ECDF). Since models cannot identify unknown data during deployment without labels, drift detection is critical for identifying potential model degradation. We adapt this concept by establishing thresholds during offline model development based on GCAPM metrics and using these to flag abnormal predictions at runtime, even in the absence of ground truth labels. 

While explainability helps identify regions that influenced prediction, it does not inherently eliminate prediction uncertainty. As illustrated in Figure 2, both high-confidence predictions and consistent attention maps can still lead to incorrect classification. Explainability, thus, is a lens into model attention, not a validation of correctness. Therefore, rather than relying solely on visual explanations, our approach introduces explainability-guided statistical reliability checks.  When predictions are statistically flagged as uncertain, the system prompts human intervention instead of issuing an unchecked result. Traditional CAM approaches are insufficient for ensuring diagnostic safety based on visual explanation alone. Our GCAPM-based safety mechanism introduces a principled framework for evaluating and trusting model outputs, improving diagnostic reliability and accountability. This is especially important for reducing risks associated with self-diagnosis and for supporting clinicians in real-world practice. 

We validate our approach using ISIC 2017 \cite{codella2018skin} and 2019 \cite{tschandl2018ham10000,codella2018skin,hernandez2024bcn20000}, two of the most widely adopted benchmarks for skin lesion diagnosis, alongside modern deep learning architectures such as CNN-based MobileNet2 \cite{sandler2018mobilenetv2} and Vision Transformer \cite{dosovitskiy2020image}, which have increasingly replaced traditional CNNs since 2020.

\section{Related Work}
\subsection{Visualizing the skin lesion classifier's diagnostic behavior}
In the study by \cite{nigar2022deep}, LIME was used to visualize model classification results which reportedly enhanced doctors' trust in the system. Similarly, \cite{hosny2024explainable} visualised model outputs via output channels to explain predictions. However, their studies did not evaluate the quality of the explanations provided by LIME, leaving it unclear whether such visualisations improved diagnostic safety. 

In \cite{esteva2017dermatologist}, saliency maps were used to visually confirm that the model focused appropriately on the lesion areas. Yet, these explanations were limited to the predicted class, and there was no analysis of how the focus point differed across other potential classes. To improve interoperability,  \cite{hryniewska2024cnn} proposed an ensemble of explainability techniques, which is useful for analyzing consistency and variation, but it lacks support for simultaneous multi-class analysis. 

A hierarchical tree-based approach in \cite{pintelas2021novel} utilized segmentation and clustering to extract texture features, aiming to present results in a human-friendly format. However, it did not evaluate how much the extracted features contributed to the model's decision. Concept Relevance Propagation \cite{achtibat2023attribution} attempts to identify not only which areas the model attends to, but also the underlying concepts those areas represent. Still, verifying the correctness of these learned concepts remains challenging and often requires expert labeling which is an inherently difficult task in medical contexts. 

Some studies used Grad-CAM saliency maps with lesion segmentation \cite{nunnari2021overlap} which resembles our approach. However, these methods typically threshold the saliency map at 0.5 for a single class, disregarding attention to alternative classes. Moreover, it is also uncertain whether this threshold is appropriate. 

Overall, previous research has mainly focused on \textit{post hoc} explanations for the predicted class and passively communicating them to clinicians. In contrast, our approach actively communicates potential model failures by visualizing attention across all relevant classes, thereby supporting both explainability and diagnostic safety.

\subsection{Safer ML using statistical data drift detection}
Runtime model monitoring and human-in-the-loop mechanisms are foundational pillars of AI safety. SafeML is a technique that uses statistical tools to monitor machine learning model performance during deployment. A key component of SafeML is the detection of data drifts, i.e., measuring the divergence between distributions of incoming data and relevant training data using statistical metrics such as ECDF-based distance measures. This allows SafeML to estimate potential performance degradation when the model encounters significantly different data from those seen during training. SafeML also brings a human into the loop, particularly when significant drifts are detected to help make safer decisions. The concept of SafeML was first introduced in \cite{aslansefat2020safeml}, and it has since evolved and been cited in the German Industry Standard for Machine Learning Uncertainty Quantification (DIN SPEC 92005) \cite{DIN92005}. 

The method was adapted for image classification tasks \cite{aslansefat2021toward} by incorporating a bootstrapping-based approach to improve the validation of distributional changes. A follow-up study applied SafeML to time-series and regression problems \cite{akram2022stadre}, where additional metrics were introduced to assess model robustness. However, a notable challenge remains in determining appropriate thresholds for drift detection. To address this, \cite{farhad2022keep} proposed an adaptive mechanism for automatic threshold selection.  

SafeML has demonstrated applicability across diverse domains, including intrusion detection \cite{aslansefat2020safeml}, autonomous driving \cite{bergler2022case,aslansefat2021toward}, robotic safety systems \cite{cho2022online,aslansefat2022safedrones}, and UAV-based wind turbine inspection \cite{kabir2022combining}. A more recent study \cite{farhad2023scope} presented a tailored version of SafeML that examines internal layers and features of neural networks to improve drift detection accuracy. This approach was further extended for Large Language Model safety under the name SafeLLM \cite{walker2024using}. 

In this paper, our distinct contribution in SafeML is an extension of the concept that integrates explainability and coverage factors to evaluate its application on skin cancer detection tasks.

\subsection{Research Questions}

Our study addresses the following research questions:

\begin{itemize}
    \item \textbf{RQ1:} Does the application of the GCAPM method improve the reliability of explainability in skin lesion classification models compared to conventional CAM-based methods?
    \item \textbf{RQ2:} How can we quantitatively evaluate the risk by comparing the quality of explainability with diagnostic performance, and what insights can be drawn from the results?
    \item \textbf{RQ3:} How can the integration of SafeML improve the detection of abnormal diagnoses in skin lesion classification models, particularly in clinical and mobile health settings?
\end{itemize}

\subsection{Contributions}

This study makes the following key contributions to enhance the performance, interpretability, and ultimately provides active safety monitoring, of skin lesion classification models: (1) GCAPM evaluation allows for more intuitive, reliable, and quantitative explanations of model predictions across all potential classes, improving on previous approaches. (2) By comparing the quality of explainability with the model's diagnostic performance, it becomes possible to establish the risk of misdiagnosis and prediction errors. (3) The integration of SafeML in this context enables proactive detection of abnormal diagnostic results and uncertainties in model predictions. 

\section{Methodology}\label{sec:methodology}
\begin{figure}
    \centering
    \includegraphics[width=1\linewidth]{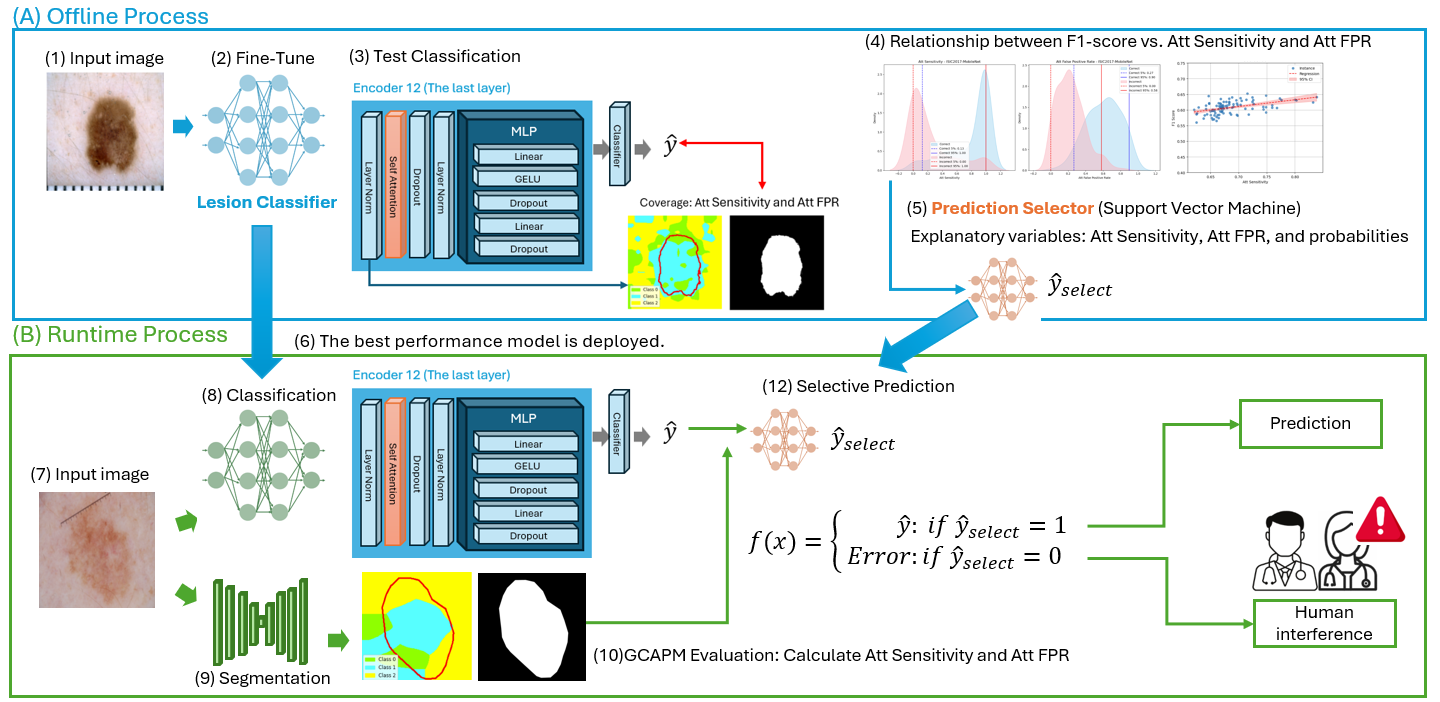}
    \caption{Safety Skin Lesion Detection Procedure: Using the knowledge gained from offline collection and accumulation of coverage estimation rates of GCAPM and segmentation, anomaly detection is performed during runtime diagnosis. When an anomaly is detected, we alter human intervention to ensure model reliability.}
    \label{fig:framework}
\end{figure}

We adopt a two-stage structure comprising an offline and a runtime process shown in Fig.\ref{fig:framework}. In the offline stage, the relationship between segmentation outputs and the qualitative metrics of explainability is analyzed. The runtime stage builds on this knowledge to detect anomalies and ensure reliable predictions.

\subsection{Global Class Activation Probabilistic Mapping (GCAPM)}\label{sec:gcapm}
The central idea in our proposal is to explain which class each pixel is associated with, rather than simply visualising pixel-level attention. While various CAM-based methods can be used for this, Grad-CAM is adopted as the base method. In Grad-CAM, the importance $\alpha^{c}_{k}$ of a target layer $A^{k}$ corresponding to the target class $c$ is computed using global average pooling of gradients as shown in Eq \ref{eq:gradcam_all}. A weighted linear combination is performed by summation and then  ReLU (Rectified Linear Unit) is applied to obtain a Grad-CAM that extracts only the scores that contribute positively to the target class.

{\small
\begin{equation}
\label{eq:gradcam_all}
\alpha^{c}_{k} = \frac{1}{Z} \sum_{i} \sum_{j} \frac{\partial y^{c}}{\partial A^{k}_{ij}}, \quad
\text{Grad-CAM} = \text{ReLU} \left( \sum_{k} \alpha^{c}_{k} A^{k} \right)
\end{equation}
}

Then, the class c that shows the highest attention at a pixel position (h,w) is selected, and its class index is stored as $C_{h,w}$ as shown in Eq. \ref{alg:gcapm}. This makes it possible to visualize which class each pixel is most strongly associated with. The GCAPM calculation process is performed according to the steps shown in Algorithm \ref{alg:gcapm}.
{\small
\begin{equation}
\label{eq:gcapm}
    C_{h,w}=\text{argmax}_{c}\mathrm{P}(c|\text{cam}(h, w))
\end{equation}}

\begin{algorithm}
\caption{GCAPM Visualization}
\label{alg:gcapm}
\begin{algorithmic}
\State \textbf{Input}: input image $x$, trained classifier $f$, \textbf{Output}: GCAPM $I$ 
\State Initialize Grad-CAM array $M$
\For {Each class $c$ in all classes of $f$}
\State Get prediction class score $y^{c}$ by $f_c(x)$
\State Calculate the gradient of the activation map $A^{k}$ to $y^{c}$ using Eq. (\ref{eq:gradcam_all})
\State Calculate Grad-CAM $m$ of $c$ by Eq. (\ref{eq:gradcam_all})
\State Apply Eq. (\ref{eq:gcapm}) to $m$.
\State Append $m$ to $M$
\EndFor
\For {pixel($i$,$j$) $i$ and $j$ are index of location of $M$}
\State Assign the class with the highest activation at $(i,j)$ to $I$
\EndFor
\State
\Return $I$
\end{algorithmic}
\end{algorithm}

\subsection{SafeML Evaluation}\label{sec:safe_evaluation}
While Mean Intersection over Union (Mean IoU) is a standard metric  for segmentation \cite{zhang2021rethinking}, it is not sufficient to capture the underestimation of attention to lesion areas. Instead we use Attribute Sensitivity (Att Sensitivity)(Eq.\ref{eq:sensitivity}) and Attribute False Positive Rate (Att FPR)(Eq.\ref{eq:fpr}). Att Sensitivity is a metric of how well the model focused on the lesion areas. Att FPR is a metric of how much the model focused on non-lesion areas by mistake. Note that in this study, we use metrics such as sensitivity and FPR, but these differ from the definitions typically used in classification tasks and measure the spatial overlap between model GCAPM and lesion regions. Therefore, the prefix ‘Att’ is used to indicate that these metrics do not explicitly indicate classification performance.
{\small
\begin{equation}
    \label{eq:sensitivity}
    \text{Att Sensitivity}=\frac{\text{True Positive}}{(\text{True Positive} + \text{False Negative})}
\end{equation}
\begin{equation}
    \label{eq:fpr}
    \text{Att FPR}=\frac{\text{False Positive}}{(\text{False Positive} + \text{True Negative})}
\end{equation}
}

\subsection{Selective Prediction}
In practical deployments, prediction labels are unavailable making it challenging to detect misclassifications. To mitigate this, we propose a selective prediction framework using a meta-classifier  (see Eq.\ref{eq:select}). 

{\small
\begin{equation}
\label{eq:select}
f(x) = \left\{ \begin{array}{cl}
\hat{y} & : \hat{y}_{select} = 1 \\
\varepsilon & : \hat{y}_{select} = 0
\end{array} \right.
\end{equation}
}

This meta-classifier takes as input the model's prediction probabilities, along with Att sensitivity, Att FPR, and performs binary classification to determine whether the prediction of the original lesion diagnosis model is accurate or not. If the meta-classifier predicts the original diagnosis correctly, the prediction is accepted. Otherwise, human intervention is triggered. This selective prediction framework helps reduce the risk of providing users with potentially incorrect diagnoses by enabling timely human oversight.

\section{Experimental Setup}
\subsection{Dataset}
\textbf{Offline Dataset:} We used the ISIC 2017 (3 classes) and the ISIC 2019 (3 classes) for training and validation.While, ISIC2019 originally contains eight classes, we selected three to simplify the validation process. These datasets are the most widely used publicly resources for skin lesion classification. To avoid duplication, we excluded overlapping samples between  ISIC2017 and ISIC2019, following the method described in \cite{cassidy2021isic}. While ISIC2017 provides segmentation annotations, ISIC2019 does not. To address this, we generated segmentation masks for ISIC2019 using DeepLabV3 with a ResNet backbone, fine-tuned on the HAM10000 \cite{tschandl2018ham10000,tschandl2020human} dataset, achieving a Mean IoU of 88\% and Pixel Accuracy of 97\%. \textbf{Runtime Dataset:} To simulate data shift conditions,  we created five versions of the original dataset by applying Gaussian blur at five intensity levels, ranging from 10\% to 50\%. As shown by the F1 scores shown in Fig.\ref{fig:box}, the performance of the model declined progressively as the level of blurring increased.

\subsection{Models}
\textbf{Lesion Classifier:} We evaluated our approach on two model architectures: the CNN-based pre-trained MobileNetV2 \cite{sandler2018mobilenetv2} and transformer-based Vision Transformer \cite{dosovitskiy2020image}. These models were trained with an initial learning rate of 1e-5. After 
 a five-epoch warm-up phase, the learning rate was gradually decreased until the final epoch. Training was run for up to 50 epochs, and the model with the lowest validation loss was selected as the best performing version. \textbf{Selective Predictor:} For the selective prediction task, we used a Support Vector Machine \cite{fan2008liblinear,chang2011libsvm,cervantes2020comprehensive} as the metalearning model. SVM was chosen for its strong generalization performance despite being a relatively simple, yet effective, weak learner.

\section{Results}
\subsection{Risk Mitigation by Improvement of Explainability}
\begin{figure}
    \centering
    \includegraphics[width=1.0\linewidth]{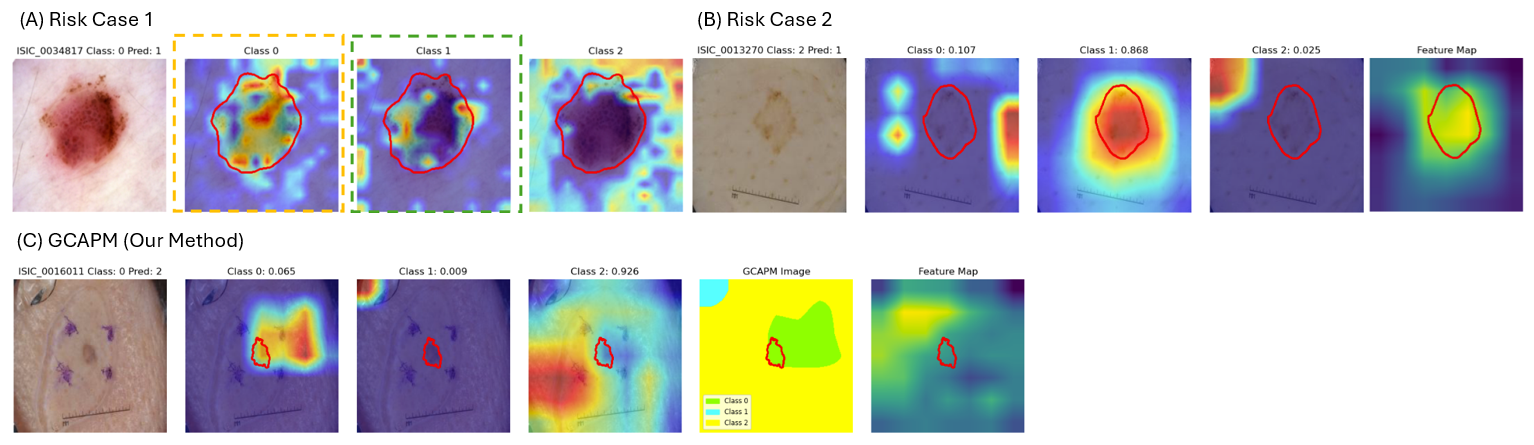}
    \caption{Risk Case 1 is Grad-CAM in classes of ISIC 2019. The red line shows the boundary of the lesion. The yellow-framed image is the actual class, and the green is the predicted class. Risk Case 2: The original image, Grad-CAM for each class, and the feature map are shown. Despite the incorrect prediction, the focus area was intensive on the lesion site. On the surface, the explanation seems reasonable. GCAPM is our method. This clearly displays potential other predictions.}
    \label{fig:cam_each_class}
\end{figure}

Results in Fig. \ref{fig:cam_each_class} demonstrate that our proposed GCAPM method effectively highlights instances where the model attends to different classes within the lesion during prediction. These nuanced patterns are often overlooked by conventional approaches that only visualise the predicted class. By enabling a deeper understanding of the model's focusing mechanisms, GCAPM enhances explainability and contributes to reducing diagnostic risk.
\vspace{-10pt}

\vspace{-40pt}

\subsection{Risk Assessment by Correlation of Coverage and Performance}
\begin{figure}[H]
    \centering

    \begin{subfigure}[b]{0.9\linewidth}
        \centering
        \includegraphics[width=0.8\linewidth]{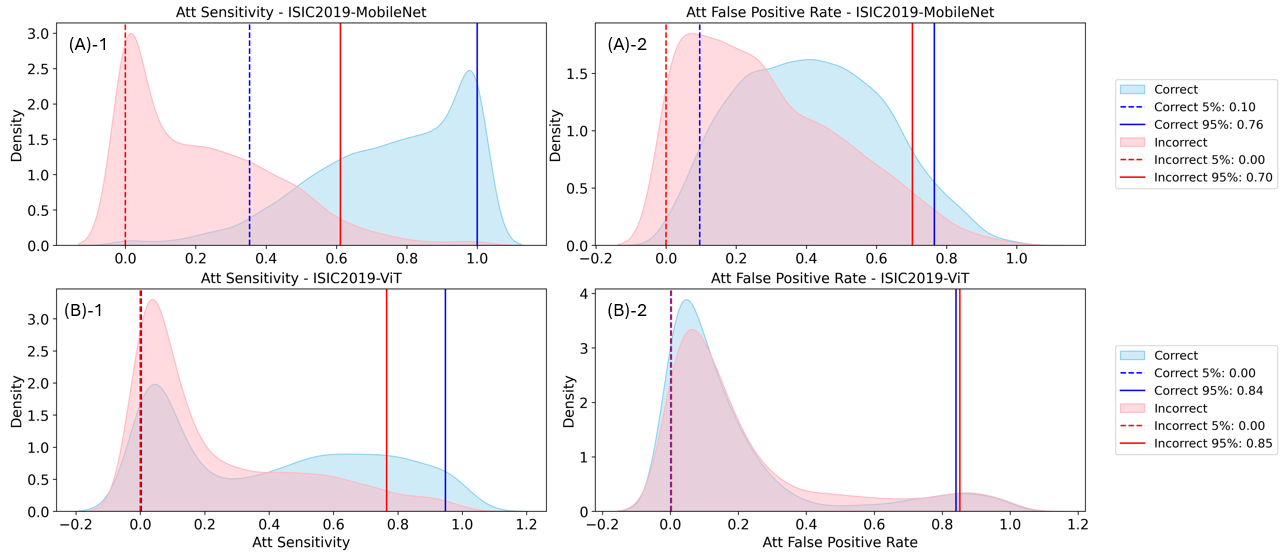}
        \caption{ISIC 2019}
        \label{fig:kde-isic2019}
    \end{subfigure}

    \begin{subfigure}[b]{0.9\linewidth}
        \centering
        \includegraphics[width=0.8\linewidth]{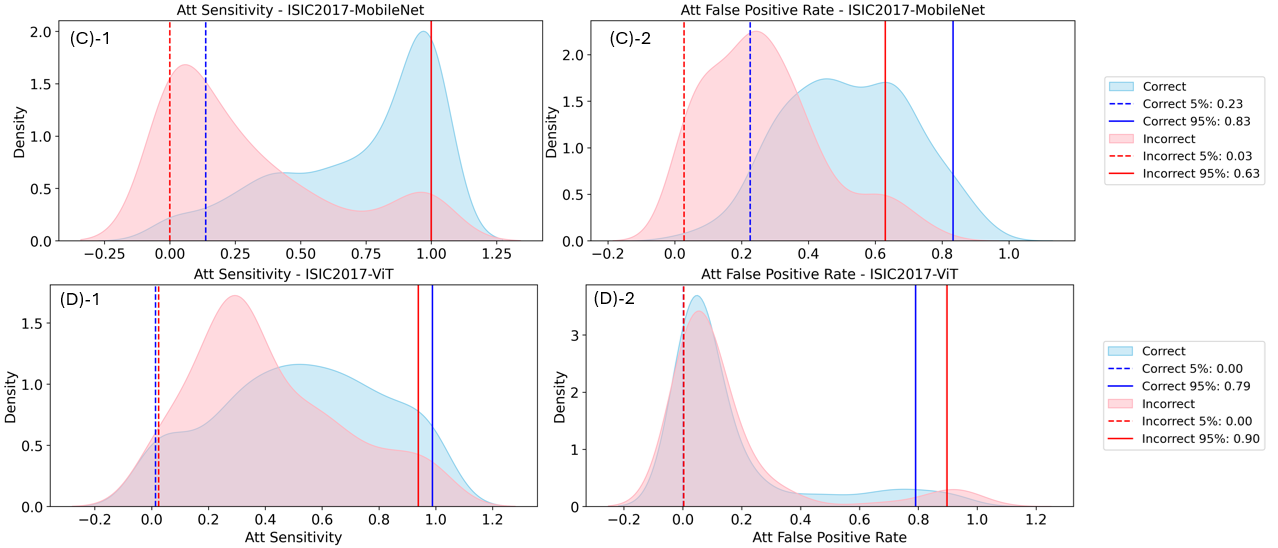}
        \caption{ISIC 2017}
        \label{fig:kde-isic2017}
    \end{subfigure}

    \caption{Density of Attributes performance:  (A) to (D) correspond to the combinations of MobileNet and Vision Transformer in ISIC2017 and ISIC2019, respectively, and show the data density of the attributes performance in each model.}
    \label{fig:kde}
\end{figure}

Fig. \ref{fig:kde} shows the distribution of data density for Att sensitivity and Att FPR. These results show clear trends in prediction behaviour. In particular, for MobileNet, inaccurate predictions tend to cluster in areas with low attribute sensitivity, while accurate predictions are more prevalent in areas of high sensitivity. In contrast, the distribution of FPR remains relatively uniform distribution for accurate predictions, indicating that as the GCAPM prediction progresses, the model's overemphasis on specific attributes decreases, leading to more stable predictions. These findings suggest a strong connection between attribute metrics and prediction accuracy.  Specifically, Att sensitivity and Att FPR can serve as supplementary indicators of trust in the model's predictions.

\vspace{-30pt}

Table \ref{tab:correlation} shows the correlation between attribute metrics and prediction performance. For the MobileNet trained on ISIC 2017, moderate positive correlations were observed between attribute metrics and prediction performance metrics, including F1 score and accuracy. A similar pattern was observed in the MobileNet-ISIC 2019 combination. On the other hand, the lesion ratio, defined as the proportion of lesion pixels in the full image, showed no significant correlation with predictive performance. This suggests that the lesion size has limited impact on model accuracy, reinforcing our hypothesis that attribute-based metrics can be reliable indicators for evaluating model trustworthiness and guiding risk mitigation.

\begin{table}
\centering
\small
\begin{tabular}{lll|c|c|c}
\hline \hline
Dataset  & Model     & Coverage        & F1 Score      & Accuracy      & Lesion ratio \\ \hline
(A) ISIC2017 & MobileNet & Att Sensitivity & \textbf{0.48} & \textbf{0.48} & -0.11        \\
         &           & Att FPR         & \textbf{0.48} & \textbf{0.50} & 0.06         \\ \hline
(B) ISIC2017 & ViT       & Att Sensitivity & 0.28          & \textbf{0.35} & -0.05        \\
         &           & Att FPR         & 0.11          & 0.08          & 0.05         \\ \hline
(C) ISIC2019 & MobileNet & Att Sensitivity & \textbf{0.66} & \textbf{0.69} & 0.08         \\ 
         &           & Att FPR         & 0.15          & 0.22          & 0.18         \\ \hline
(D) ISIC2019 & ViT       & Att Sensitivity & -0.14         & -0.12         & 0.07         \\
         &           & Att FPR         & 0.00          & 0.03          & 0.04         \\ \hline \hline
\end{tabular}
\caption{Correlation Between Attribute Metrics and Predictive Performance}
\label{tab:correlation}
\end{table}

\begin{figure}[H]
    \centering
    \includegraphics[width=0.9\linewidth]{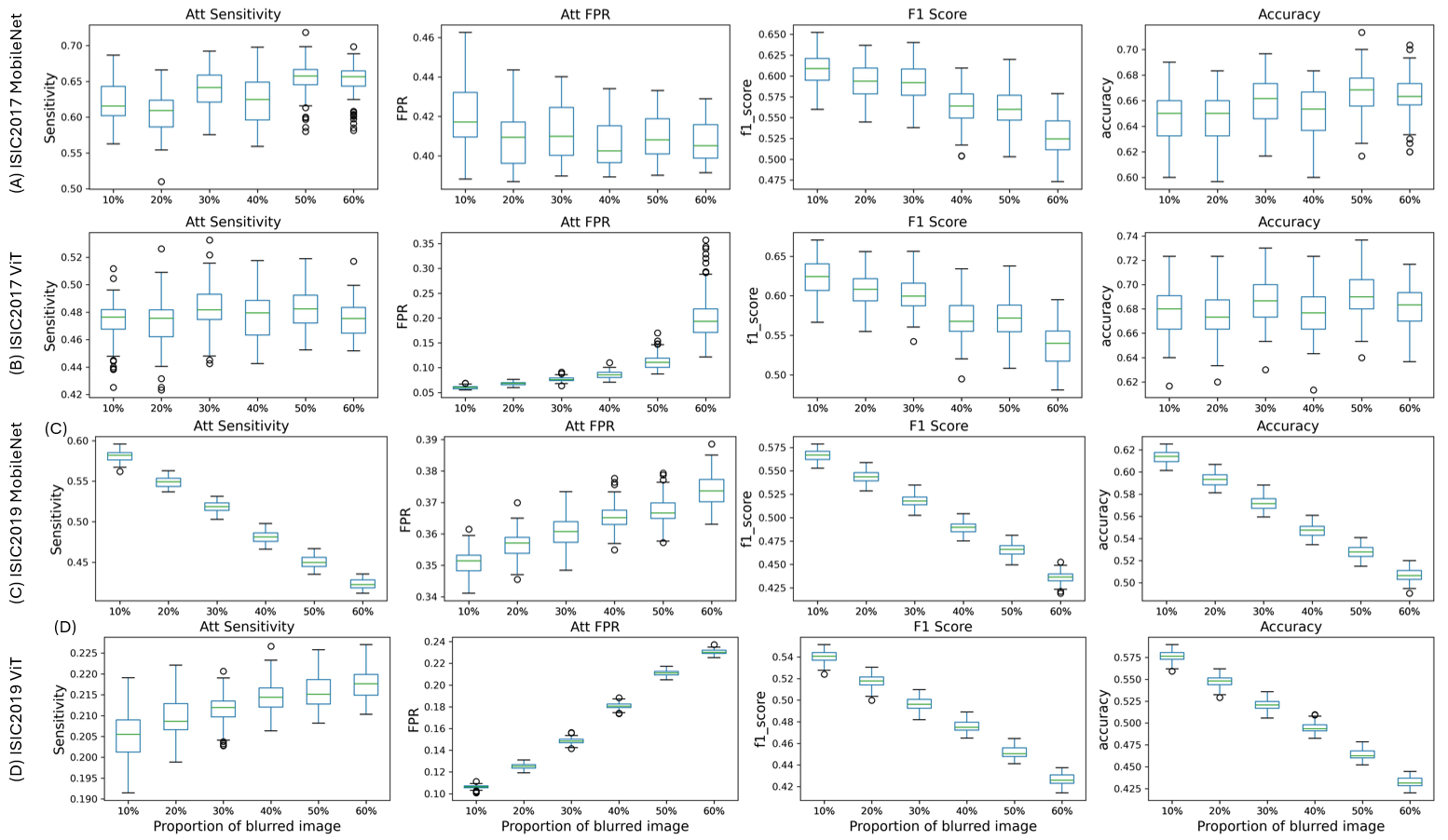}
    \caption{Attributes and Predictive Performance each Data: (A) shows the MobileNet model from ISIC 2017, (B) shows the Vision Transformer from ISIC 2017, (C) shows the MobileNet from ISIC 2019, and (D) shows the changes in attribute and prediction performance when using the Vision Transformer from ISIC 2019.}
    \label{fig:box}
\end{figure}

\subsection{Improvement Abnormal Diagnosis Detection in Runtime}
\begin{figure}
    \centering
    \includegraphics[width=1\linewidth]{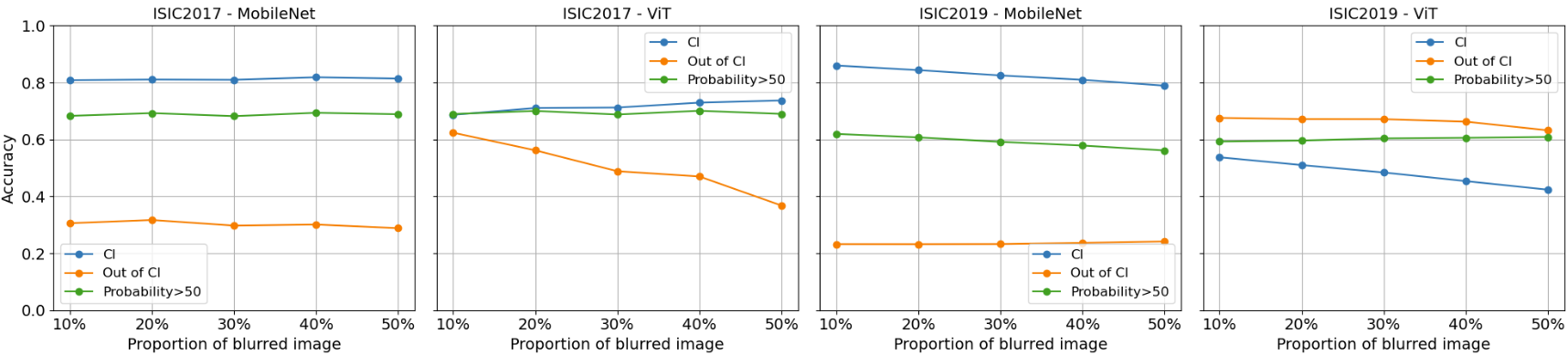}
    \caption{Prediction Accuracy in CI, Out of CI, and Probability}
    \label{fig:ci}
\end{figure}

Fig. \ref{fig:ci} presents the results of evaluating prediction accuracy both within and outside the confidence interval (CI) (Fig.\ref{fig:kde}) of the attribute metrics and for data with a prediction probability of 50\% or more presented in Figure 5. The experimental findings show that the lesion classifier achieves high diagnostic accuracy within the CI. Specifically, MobileNet yielded accurate predictions for around 80\% of the data within the CI, while the Vision Transformer (ViT) on the ISIC 2017 achieved correct predictions for approximately 70\% of the data. In contrast, prediction accuracy outside the CI dropped to around 30\%, indicating a substantial decline in reliability.

Furthermore, the prediction accuracy within the CI was consistently higher than that of data with prediction probabilities exceeding 50\%. This suggests that CI that is derived from attribute metrics offers a more precise and reliable indication of prediction certainty than probability scores alone.

\begin{table}[]
\centering
\small
\begin{tabular}{ll|rr|rr|rr|rr|rr}
\hline \hline
\multicolumn{1}{c}{\multirow{2}{*}{Dataset}} & \multicolumn{1}{c|}{\multirow{2}{*}{Model}} & \multicolumn{2}{c|}{10\%}              & \multicolumn{2}{c|}{20\%} & \multicolumn{2}{c|}{30\%} & \multicolumn{2}{c|}{40\%} & \multicolumn{2}{c}{50\%} \\
\multicolumn{1}{c}{} & \multicolumn{1}{c|}{} & \multicolumn{1}{l}{Acc} & \multicolumn{1}{l|}{Inacc} & \multicolumn{1}{l}{Acc} & \multicolumn{1}{l|}{Inacc} & \multicolumn{1}{l}{Acc} & \multicolumn{1}{l|}{Inacc} & \multicolumn{1}{l}{Acc} & \multicolumn{1}{l|}{Inacc} & \multicolumn{1}{l}{Acc} & \multicolumn{1}{l}{Inacc} \\ \hline
(A) ISIC2017 & MobileNet & 0.91 & 0.80 & 0.91 & 0.77 & 0.89 & 0.77 & 0.88 & 0.76 & 0.86 & 0.74                      \\
(B) ISIC2017 & ViT & 0.90 & 0.68 & 0.87 & 0.66 & 0.82 & 0.66 & 0.82 & 0.68 & 0.84 & 0.75                      \\
(C) ISIC2019 & MobileNet & 0.94 & 0.85 & 0.93 & 0.85 & 0.92 & 0.85 & 0.90 & 0.84 & 0.88 & 0.83                      \\
(D) ISIC2019 & ViT & 0.96 & 0.89 & 0.96 & 0.87 & 0.96 & 0.85 & 0.96 & 0.84  & 0.95 & 0.83                      \\ \hline \hline
\end{tabular}
\caption{Improvement of Accuracy by Runtime Selective Prediction}
\label{tab:selective}
\end{table}

\vspace{-20pt}

Table \ref{tab:selective} shows the results of applying  a runtime selective prediction strategy. This experiment tested whether a selective predictor could reliably determine the accuracy of the lesion classifier's predictions under increasingly challenging conditions, including 10\% to 50\% artificially produced contamination to the data, as shown in Fig. \ref{fig:box}. 

Even under these challenging conditions, the selective predictor idenitifed correct predictions with approximately 90\% accuracy across most settings. Moreover, it successfully flagged inaccurate predictions with over 75\% accuracy in all cases except Pattern (B). These results confirm the effectiveness of the proposed framework in proactively detecting and avoiding high risk misdiagnoses.  They show a potential for this method to significantly improve diagnostic safety by enabling the system to prevent predictions likely ot be incorrect, reducing the risk of misdiagnosis  by a very large margin in most of our  experiments.

\section{Discussion and Future Work}
\subsection{Discussion}
The experimental findings demonstrate that GCAPM significantly contributed to explainability, i.e. clarifying the reason for diagnosis, and reducing the risk of misdiagnosis. By visualizing class-wise attention lesion areas, even in challenging or ambiguous cases that are often overlooked with conventional methods, GCAPM helped clarify the reasoning behind predictions.

Moreover, accurate predictions tended to concentrate in areas with high Att Sensitivity, while incorrect predictions tended to occur in regions with low Att Sensitivity. Att FPR remained relatively uniform reflecting the model's ability to converge towards the correct class while reducing excessive dependence on any particular class. These results suggest that Att sensitivity and Att FPR, as measured in GCAPM, are effective indicators of prediction reliability.

Furthermore, data within the confidence interval CI derived from attribute metrics consistently showed higher diagnostic accuracy than data selected based solely on prediction probability. Additionally, under simulated conditions of degraded model performance  (with 10-50\% data contamination), the selective predictor maintained approximately 90\% accuracy in distinguishing correct from incorrect predictions. In misdiagnosed cases, risk was successfully detected with 75\% accuracy, excluding pattern (B) in Table \ref{tab:selective}.

Collectively, these findings indicate  that our safety monitoring framework, introducing GCAPM and selective predictors inspired by SafeML, has the potential to significantly improve the performance of diagnostic support by proactively avoiding predictions with a high risk of misdiagnosis.
\vspace{-10pt}

\subsection{Future Work}
In this study, we simulated data drift by adding random noise. Future work  will focus on evaluating fairness and generaliztion by analysing the impact of skin tone variations on diagnoses. While public datasets used in this study contain mainly light skin tones,  realistic environments are likely to encounter more diverse skin tone populations. Assessing the framework's adaptability and fairness under these conditions is a key next step. Moreover, we plan to explore applicability in broader AI domains, including Generative AI models like Large Language Models and Large Vision Models.

\vspace{-10pt}

\subsection{Research Limitations}

This study utilised a CAM method compatible with both ViT and MobileNet. GCAPM was computed probabilistically across all classes. While ViT has the potential to extract richer and more meaningful features through its self-attention mechanism,  the explainability of attention layers is not always intuitive or interpretable. Although advanced methods such as Attention Rollout \cite{abnar2020quantifying}, Attention Flow \cite{abnar2020quantifying}, and Deep Taylor Decomposition \cite{chefer2021transformer} have been proposed, explanability research specific to ViT remains at early stages. Further work is needed to develop methods that offer transparent and reliable interpretation of ViT structures. Additionally, the proposed method assumes access to the internal structure of the model, potentially limiting applicability to commercial and proprietary systems. In such cases, model-agnostic approaches like SMILE \cite{aslansefat2023explaining} can be considered.

\section{Conclusion}
This study addressed two issues associated with deploying DL in medical diagnosis: (1) the risk that a false prediction may appear trustworthy due to misleading single-class explanability visualization, and (2) the risk related to the inherent uncertainty in deployment scenarios where ground-truth labels are unavailable. 
The former risk was avoided using GCAPM, while the latter uncertainty was reduced by combining GCAPM with a selective predictor. Our results demonstrate that the proposed system contributes to enabling safer and more reliable use of DL in medical settings, with limitations noted and further work outlined.

%
%

\begin{credits}
\subsubsection{\ackname} 
The authors would like to thank the Data Science, Artificial Intelligence, and Modelling (DAIM) Institute at the University of Hull for their support. Furthermore, the authors extend their heartfelt gratitude to Dr. Jun-ya Norimatsu at ALINEAR Corp. for technical advice with the experiments.

\end{credits}

\bibliographystyle{splncs04}
\bibliography{reference}

\end{document}